\documentclass[conference]{IEEEtran}

\IEEEoverridecommandlockouts
\usepackage{graphicx}
\usepackage{xspace}
\usepackage{svg}
\usepackage{circuitikz}
\usepackage[utf8]{inputenc}
\usepackage{tabularx}
\usepackage{booktabs}
\usepackage[format=default,font=footnotesize]{caption}
\usepackage[flushleft]{threeparttable}
\usepackage[T1]{fontenc}
\usepackage[numbers,sort&compress]{natbib}
\usepackage{amsmath,amsfonts}
\usepackage{amssymb}
\usepackage{color,soul}
\usepackage{makecell}
\usepackage{balance}
\usepackage{xcolor}
\usepackage{dblfloatfix}
\usepackage{fancyhdr}  
\usepackage[ruled, vlined]{algorithm2e}
\usepackage[a4paper,total={184mm,239mm}]{geometry}
\def\BibTeX{{\rm B\kern-.05em{\sc i\kern-.025em b}\kern-.08em
    T\kern-.1667em\lower.7ex\hbox{E}\kern-.125emX}}

\usepackage{babel}
\usepackage{algorithmicx}
\usepackage[colorlinks=true,urlcolor=blue,citecolor=blue,linkcolor=blue]{hyperref}
\usepackage{multirow}
\usepackage[sets,nn,operations,colors]{cora-macs}

\usepackage{xcolor}
\usepackage{lipsum}
\setlipsum{%
    par-before = \begingroup\color{lightgray},
    par-after = \endgroup,
    sentence-before = \begingroup\color{lightgray},
    sentence-after = \endgroup
}

\usepackage{amsthm}
\usepackage{cleveref}
\usepackage{graphicx}
\usepackage{amsfonts}
\usepackage{wasysym}

\newtheorem{proposition}{Proposition}

\newtheorem{definition}{Definition}

\crefname{section}{Sec.}{Sec.}
\crefname{subsection}{Sec.}{Sec.}
\crefname{figure}{Fig.}{Fig.}
\crefname{algorithm}{Alg.}{Alg.}
\crefname{table}{Tab.}{Tab.}
\crefname{example}{Ex.}{Ex.}
\crefname{definition}{Def.}{Def.}
\crefname{proposition}{Prop.}{Prop.}
\crefname{corollary}{Cor.}{Cor.}
\crefname{theorem}{Thm.}{Thm.}
\crefname{lemma}{Lemma}{Lemmas}
\crefname{appendix}{Appendix}{Appendix}

\Crefname{section}{Sec.}{Sec.}
\Crefname{subsection}{Sec.}{Sec.}
\Crefname{figure}{Fig.}{Fig.}
\Crefname{algorithm}{Alg.}{Alg.}
\Crefname{table}{Tab.}{Tab.}
\Crefname{example}{Ex.}{Ex.}
\Crefname{definition}{Def.}{Def.}
\Crefname{proposition}{Prop.}{Prop.}
\Crefname{theorem}{Thm.}{Thm.}
\Crefname{corollary}{Cor.}{Cor.}
\Crefname{lemma}{Lemma}{Lemmas}
\Crefname{appendix}{Appendix}{Appendix}

\usepackage{nicefrac}

\usepackage{tikzscale}
\usepackage{pgfplots}
\usepackage{mathtools}
\pgfplotsset{compat=1.18}
\usepgfplotslibrary{external}
\tikzset{every picture/.style={line width=0.5pt}}
\usepgfplotslibrary{groupplots}
\usetikzlibrary{pgfplots.groupplots}
\usetikzlibrary{matrix}
\usetikzlibrary{backgrounds}

\usepackage{mathtools,stackengine}
\stackMath
\newcommand{\stackEq}[2]{%
    \setbox0=\hbox{${}\mathrel{\stackon[-1pt]{#2}{\scriptstyle #1\strut}}{}$}
    \xdef\tmpwd{\dimexpr\the\wd0\relax}
    \kern.5\tmpwd\mathclap{\box0}&\kern.5\tmpwd
}

% ...
\usepackage{caption, subcaption}
\usepackage{duckuments}

\usepackage{xparse}
\usepackage{listofitems}
\newcommand{\confusionmatrix}[5]{
    \tikzset{
        cell/.style={rectangle, minimum width=#5, minimum height=#5, draw, font=\footnotesize},
        label/.style={font=\small},
        axis_label/.style={font=\bfseries}
    }

    \pgfmathsetmacro{\scale}{#5/1cm}

    \readlist\classlist{#2}
    \readlist\correctlist{#3}
    \readlist\unknownlist{#4}

    \foreach \i in {1,...,#1} {
        \foreach \j in {1,...,#1} {
            \node[cell] (cell-\i-\j) at ({\scale*(\j-1)},-{\scale*(\i-1)}) {};
        }
    }

    \foreachitem \class \in \classlist[]{
        \node[label, anchor=south, rotate=90] at ({-0.5*\scale},-{\scale*(\classcnt-1)}) {$\mathstrut \text{\class}$};
        \node[label, anchor=south] at ({\scale*(\classcnt-1)},{0.5*\scale}) {$\mathstrut \text{\class}$};
    }
    \foreach \i in {1,...,#1} {
        \foreach \j in {1,...,#1} {
            \pgfmathsetmacro{\index}{int((\i-1)*#1+\j)}
            \edef\correct{\correctlist[\index]}
            \edef\unknown{\unknownlist[\index]}
            \pgfmathsetmacro{\total}{\correct+\unknown}
            \pgfmathtruncatemacro{\totalint}{\total}
            \ifnum\totalint>0
            \pgfmathsetmacro{\correctheight}{\correct/\total}
            \pgfmathsetmacro{\unknownheight}{\unknown/\total}
            \fill[CORAcolor3!50] (cell-\i-\j.south east) rectangle ($(cell-\i-\j.north west)!{\correctheight}!(cell-\i-\j.south west)$);
            \fill[CORAcolor5!50] ($(cell-\i-\j.north west)!{\correctheight}!(cell-\i-\j.south west)$) rectangle (cell-\i-\j.north east);
            \node[cell] at (cell-\i-\j.center) {$\correct|\unknown$};
            \else
                \node at (cell-\i-\j.center) {};
            \fi
        }
    }

    \node[axis_label, anchor=center] at ({\scale*(#1/2-0.5)},{-(#1)*\scale}) {True class};
    \node[axis_label, rotate=90, anchor=south] at ({-1.25cm},{-\scale*(#1/2-0.5)}) {Predicted class};
}

\newcommand{\ReLU}{\text{\normalfont ReLU}}

\makeatletter
\let\ps@plain\ps@empty
\let\ps@headings\ps@empty
\makeatother

\begin{document}
\title{Formally Verifying Analog Neural Networks Under Process Variations Using Polynomial Zonotopes}
% uncomment for anonymous submission
% \author{\IEEEauthorblockN{Anonymous Authors}}

% actual author block
 \author{
      \IEEEauthorblockN{Yasmine Abu-Haeyeh$^{\dagger1}$,  Tobias Ladner$^{\dagger2}$, Matthias Althoff$^2$, Lars Hedrich$^1$}
     \IEEEauthorblockA{
         $^1$ Goethe University Frankfurt, Germany \\
         \texttt{\{abu-haeyeh,hedrich\}@em.uni-frankfurt.de}
     }
      \IEEEauthorblockA{
        $^2$ Technical University of Munich, Germany \\
        \texttt{\{tobias.ladner,althoff\}@tum.de}
      }
 \vspace{-0.7cm}
 }

% make title
\maketitle

% footnote equal contribution!
\begingroup\renewcommand\thefootnote{$^\dagger$}
  \footnotetext{Equal contribution, sorted alphabetically.}
\endgroup

% IEEE copyright notice for camera-ready (IEEE Xplore proceedings), FDL 2026.
% Placed as an unmarked footnote because the page-style overrides above disable
% the page-foot hook that \IEEEpubid relies on.
% \begingroup\renewcommand\thefootnote{}%
%   \footnotetext{979-8-3195-3997-7/26/\$31.00~\copyright2026 IEEE}%
%   \addtocounter{footnote}{-1}%
% \endgroup

% TODO: is this required?
% for having page numbers in 
%\todo{could be removed for publication}
% \thispagestyle{empty}
% \pagestyle{empty}
\thispagestyle{plain}
\pagestyle{plain}
%\thispagestyle{fancy}
%\pagestyle{fancy}
%%\todo{Programmable?}
% remove of toplevel rule: 
% \renewcommand{\headrulewidth}{0pt}
% verschieben der Seitenzahlen: 
\cfoot{\vspace*{-0.05\baselineskip}\thepage}

%\vspace{-1cm}

% abstract

\begin{abstract}    
    Analog neural networks are gaining attention due to their efficiency in terms of power consumption and processing speed. 
    However, since analog neural networks are implemented as physical circuits, they are highly sensitive to manufacturing process variations,
    which can cause large deviations from the nominal model.    
    We present a polynomial-based model that resembles the performance of the neuron circuit under process variations. 
    This model is formally verified via reachability analysis using polynomial zonotopes,
    thus avoiding conventional, time-consuming Monte Carlo simulations.
    We evaluate our proposed verification approach on three different datasets and on fully-connected and convolutional analog neural networks.
    Our experimental results confirm the effectiveness of our verification approach by reducing the verification time from up to a day to seconds while enclosing up to $99 \%$ of the variation samples.
\end{abstract}

\begin{IEEEkeywords}
    Analog neural networks, formal verification, set-based computing, energy-efficient computing, AI hardware.
\end{IEEEkeywords}

% sections

\section{Introduction}\label{sec:introduction}

The growing demand for artificial neural networks has driven the need for low-power hardware solutions, such as FPGA accelerators, in-memory computing, and CMOS-based inference circuits~\cite{dhilleswararao2022efficient, datar2024promise}.
This facilitates the deployment of neural networks in Internet of Things (IoT) devices, such as sensors, mobile devices, and wearable technology, where power consumption is a critical concern~\cite{yan2023polythrottle, gao2025overview, tain2025j3dai}.
Analog inference hardware demonstrates efficiency in terms of processing speed, energy, and chip area.
However, the design and verification of analog devices are challenging due to their vulnerability to perturbations such as noise, mismatch, and process variations~\cite{li2007statistical,yan2022computing, xiao2023accuracy}.
The functionality of analog neural networks must be validated under these variations to ensure safety, which is usually performed by running time-consuming Monte Carlo simulations~\cite{gencer2021design}.
In this work, we formally verify the characteristics of analog neural networks influenced by varying process parameters.

\subsection{Related Work}

Modeling and analysis of the influence of varying process parameters on analog circuits have been investigated extensively~\cite{csandru2021modeling, narayanan2010formal, lyu2023machine, kim2025glova}.
However, recent advances in the design and implementation of analog neural networks have opened up new challenges that need to be further addressed:
During semiconductor device fabrication, small variations---such as gate width, threshold voltage, channel length, and oxide thickness---naturally occur~\cite{huff2021important}.
It is well known that small variations can significantly impact the accuracy of conventionally implemented neural networks~\cite{goodfellow2015explaining},
and similar effects have also been observed for process variations on analog neural networks~\cite{yan2022computing,xiao2023accuracy,afroz2024sensitivity,jia2018calibrating}.
Hence, modeling the behavior of analog inference circuits under these variations is crucial to ensure the safety of these circuits and improve their design and robustness~\cite{ladner2025analog}.
The emerging modeling methodologies either rely on device physics or machine learning engines~\cite{li2024overview}.
Since developing accurate physics-based models is a challenging and time-consuming task, many works utilize machine learning for modeling analog circuits under process variations~\cite{lyu2023machine}.
While such models provide the capability of identifying the sources of variability in the circuit,
they are usually single-transistor models, which are insufficient to reflect the full system behavior and therefore cannot be extended to a system-level modeling approach.
Other available analog neural network implementations employ Monte Carlo simulations to analyze the behavior of specific circuits under process variations~\cite{gencer2021design, nagele2023analog}.
However, the simulations have to be rerun for each network architecture, which can take days as up to $100,000$ transistors have to be simulated.

Recently, it was shown that system-level verification of analog neural networks under device mismatch can be realized using formal neural network verification~\cite{ladner2025analog}.
Neural network verification can provide formal guarantees of the prediction of a neural network under perturbations, where usually conventionally implemented neural networks are considered~\cite{kaulen20256th,manzanas2024arch}.
These guarantees are obtained by rigorous mathematical proofs that enable reasoning about the safety of a neural network under perturbation~\cite{huang2017safety}.
Unfortunately, these proofs are usually difficult to obtain for large neural networks as it has been shown that verifying certain properties is NP-hard~\cite{katz2017reluplex}.
Thus, modern verifiers introduce relaxations to remain computationally feasible,
where both optimization-based approaches~\cite{zhang2018efficient, katz2019marabou, henriksen2020efficient, singh2019abstract} and approaches based on reachability analysis~\cite{gehr2018ai2, singh2018fast,lopez2023nnv,ladner2023automatic} are used.

\subsection{Contributions}

To summarize, our contributions are:
\begin{itemize}
    \item An optimized polynomial-based parametric model for the analog neuron performance under process variations.
    \item The system-level verification of the model using reachability analysis.
    \item The scalability is demonstrated on various network architectures, including fully-connected and convolutional networks,
    and on benchmark classification tasks such as MNIST.
\end{itemize}

\section{Preliminaries}\label{sec:preliminaries}

\subsection{Notation}

We denote scalars and vectors by lowercase letters, matrices by uppercase letters, and sets by calligraphic letters.
 The $i$-th element of a vector $v\in\R^n$ is written as $v_{(i)}$.
The element in the $i$-th row and $j$-th column of a matrix $A\in\R^{n\times m}$ is written as $A_{(i,j)}$,
 the entire $i$-th row and $j$-th column are written as $A_{(i,\cdot)}$ and $A_{(\cdot,j)}$, respectively.
This is analogously used for tensors.
% The concatenation of $A$ with a matrix $B\in\R^{n\times o}$ is denoted by $[A\ B]\in\R^{n\times (m+o)}$.
% The empty matrix is written as~$[\ ]$.
% We denote with $I_n$ the identity matrix of dimension $n\in\mathbb{N}$.
% We denote the element-wise multiplication of two vectors by $\odot$.
% The bold symbol $\mathbf{1}$ refers to a matrix with all ones of proper dimensions.
% The symbol~$\mathbf{1}$ refers to the matrix with all ones of proper dimensions.
 Given $n\in\mathbb{N}$, we use the shorthand notation $[n]=\left\{ 1,\ldots,n \right\}$.
% % Given $a \leq b\in\mathbb{N}_0$, then $[b]_{a}=\left\{ a,a+1,\ldots,b \right\}$ and $[b] = [b]_{1}$.
% % The cardinality of a discrete set $\mathcal{D}$ is denoted by $|\mathcal{D}|$.
% % Let $\mathcal{D}\subseteq [n]$, then $A_{(\mathcal{D}, \cdot)}$ denotes all rows $i\in\mathcal{D}$ in lexicographic order;
% % this is used analogously for columns.
% % We denote the power set of $\mathcal{D}$ by $2^{\mathcal{D}}$.
Let $\mathcal{S}\subset\R^n$ be a set
% % then $\mathcal{S}_{(i)}$ is its projection on the $i$-th dimension.
and $f\colon \R^n\rightarrow \R^m$ be a function,
 then $f(\mathcal{S}) = \left\{f(x)\ \middle|\ x \in \mathcal{S}\right\}$.
% Given two sets $\mathcal{S}_1,\mathcal{S}_2$,
% then the Minkowski sum is denoted by $\mathcal{S}_1 \oplus \mathcal{S}_2 = \{ s_1 + s_2\ |\ s_1 \in \mathcal{S}_1,  s_2 \in \mathcal{S}_2 \}$.
% % The Cartesian product is written as $\mathcal{S}_1 \times \mathcal{S}_2 = \{ \cmatrix{s_1 & s_2 }^\top |\ s_1 \in \mathcal{S}_1,
% % s_2 \in \mathcal{S}_2 \}$.
 An interval with bounds $a,\,b\in\R^n$ is denoted by $[a,\,b]$, where $a\leq b$ holds element-wise.

\subsection{Neural Networks}

We initially consider fully-connected neural networks with rectified linear unit (ReLU) activation, and later extend our approach to convolutional neural networks~\cite[Sec.~5.5.6]{bishop2006pattern}.
\begin{definition}[ReLU Neural Network~{\cite[Sec.~5.1]{bishop2006pattern}}]
    \label{def:neural-networks}
    Given an input $\nnInput\in\R^{\numNeurons_0}$ and an output $\nnOutput\in\R^{\numNeurons_\numLayers}$,
    a neural network $y=\NN(x)$ with $\numLayers\in\N$ layers can be formulated as
    \begin{align*}
        \begin{split}
            \nnHidden_0 &= \nnInput\text{,} \\
            \nnHidden_k &= \nnLayer{k}{\nnHidden_{k-1}} = \ReLU(W_k \nnHidden_{k-1} + b_k), \quad k\in[\numLayers], \\
            \nnOutput &= \nnHidden_\numLayers\text{,}
        \end{split}
    \end{align*}
    with outputs of hidden layers $\nnHidden_k\in\R^{\numNeurons_k}$, layer $L_k\colon \R^{\numNeurons_{k-1}}\to\R^{\numNeurons_k}$, weights $W_k\in\R^{\numNeurons_k\times\numNeurons_{k-1}}$, and bias $b_k\in\R^{\numNeurons_k}$.
\end{definition}

\subsection{Set-Based Computing}
\label{sec:set-based-computing}

We bound variations using continuous sets.
To efficiently represent them in high-dimensional spaces, 
we use (matrix) polynomial zonotopes~\cite{kochdumper2020sparse,ladner2025formal} as a set representation.
\begin{definition}[Matrix Polynomial Zonotope~{\cite[Sec.~3]{ladner2025formal}}]
    \label[definition]{def:pZ-mat}
    Given an offset $C\in\R^{n\times m}$,
    dependent generators $G \in \R^{n\times m \times h}$,
    independent generators $G_I \in\R^{n\times m\times q}$,
    and an exponent matrix $E\in \mathbb{N}_0^{p\times h}$,
    a matrix polynomial zonotope $\PZ = \shortPZ{C}{G}{G_I}{E}\subset\R^{n\times m}$ is defined as
    \begin{align*}
        \PZ &\coloneqq \left\{ C + \sum_{i=1}^h \left(\prod_{k=1}^p \alpha_k^{E_{(k,i)}}\right) G_{(\cdot,\cdot, i)} + \sum_{j=1}^{q} \beta_j G_{I(\cdot,\cdot, j)}\ \right|\\ %
        &\qquad\qquad\qquad\qquad\qquad \alpha_k, \beta_j \in \left[-1,1\right] \bigg\}.
    \end{align*}
\end{definition}
Many of the required operations to verify analog neural networks can be computed efficiently using polynomial zonotopes.
We discuss them only briefly here due to space limitations, and refer interested readers to the example in \cite[Appendix~A]{ladner2025formal}.
Given a matrix polynomial zonotope $\PZ_1=\shortPZ{C_1}{G_1}{G_{I,1}}{E_1}$, a matrix $A\in\R^{k\times n}$, and a matrix $B\in\R^{k\times m}$,
an affine map~\cite[Eq.~6a]{ladner2025formal} is computed by
\begin{align}
    \label{eq:pz-affine-map}
    \begin{split}
        A\PZ_1+B &= \left\{ A X_1 + B\ \middle|\ X_1\in\PZ_1 \right\} \\
        &= \shortPZ{A C_1 + B}{A G_1}{A G_{I,1}}{E_1}.
    \end{split}
\end{align}
Given a second matrix polynomial zonotope $\PZ_2$ with matching dimensions,
one can also efficiently compute their sum~\cite[Eq.~5]{ladner2025formal}
\begin{equation}
    \label{eq:pz-sum}
    \PZ_1\oplus\PZ_2 = \left\{ X_1 + X_2\ \middle|\ X_1\in\PZ_1,\, X_2\in\PZ_2\right\},
\end{equation} 
and their multiplication~\cite[Lemma~1]{ladner2025formal}
\begin{equation}
    \label{eq:pz-multiplication}
    \PZ_1 \PZ_2 = \left\{ X_1 X_2\ \middle|\ X_1\in\PZ_1,\, X_2\in\PZ_2\right\}.
\end{equation}

\subsection{Formal Verification of Neural Networks}
\label{sec:verification-neural-network}

We enclose all possible outputs of a neural network for a given input set $\nnInputSet\subset\R^{\numNeurons_0}$ by
\begin{align}
    \begin{split}
        \nnHiddenSetExact_0 &= \nnInputSet, \\
        \nnHiddenSetExact_k &= \nnLayer{k}{\nnHiddenSetExact_{k-1}}, \quad\ k\in[\numLayers], \\
        \nnOutputSetExact &= \nnHiddenSetExact_\numLayers.
    \end{split}
\end{align}
It is generally computationally infeasible to obtain these true sets (indicated by $\square^*$)~\cite{katz2017reluplex}
so that the output of each layer is enclosed:
\begin{proposition}[Neural Network Enclosure~{\cite[Sec.~3]{singh2018fast}}]
    \label{prop:nnv-enclosure}
    Given an input set $\nnInputSet\subset\R^{\numNeurons_0}$ to a neural network $\NN$,
    we obtain an enclosure of the output set $\nnOutputSet = \opEnclose{\NN}{\nnInputSet}\supseteq\nnOutputSetExact$ by iteratively enclosing the output of each layer:
    \begin{align*}
        \nnHiddenSet_k = \opEnclose{L_k}{\nnHiddenSet_{k-1}} \supseteq \nnHiddenSetExact_k\text{,} \quad k\in[\numLayers]\text{,}
    \end{align*}
    with $\nnHiddenSet_0 = \nnInputSet$ and $\nnOutputSet = \nnHiddenSet_{\numLayers}$.
\end{proposition}
We utilize the enclosure $\nnOutputSet\supseteq\nnOutputSetExact$ of the output set to verify that unsafe outputs are impossible:
\begin{equation}
    \label{eq:problem-statement}
    \nnOutputSet\cap\unsafeSet = \emptyset \implies \nnOutputSetExact\cap\unsafeSet = \emptyset,
\end{equation}
where $\unsafeSet\subset\R^{\numNeurons_\numLayers}$ defines a set of unsafe outputs.
Many common verification tasks such as image classification can be formulated this way.

\begin{figure}[t]
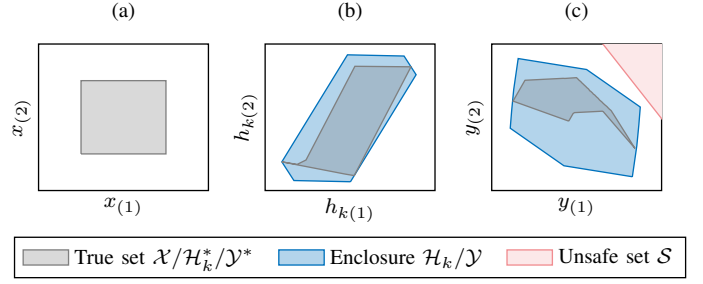

    \centering
    \tikzsetnextfilename{./figures/nnv/nnv-example}%
    \definecolor{mycolor1}{rgb}{0.00000,0.44700,0.74100}%
\definecolor{mycolor2}{rgb}{0.85000,0.32500,0.09800}%
\definecolor{mycolor3}{rgb}{0.92900,0.69400,0.12500}%
\definecolor{mycolor4}{rgb}{0.49400,0.18400,0.55600}%
\begin{tikzpicture}
\footnotesize
\pgfplotsset{
    plotstyle1/.style={area legend, draw=CORAcolorBlue, fill=CORAcolorBlue, fill opacity=0.3, forget plot},
    plotstyle2/.style={only marks, mark=*, mark options={}, mark size=0.000pt, legend image post style={mark size=1pt}, draw=black, forget plot},
    plotstyle3/.style={area legend, draw=gray, fill=gray, fill opacity=0.3},
    plotstyle4/.style={area legend, draw=CORAcolorUnsafe, fill=CORAcolorUnsafeLight, forget plot},
}
\def\rows{1}
\def\cols{3}
\def\horzsep{0.75cm}
\def\basepath{./figures/nnv/}

\begin{groupplot}[%
group style={rows = \rows, columns = \cols, horizontal sep = \horzsep},
scale only axis,
width=1/\cols*\linewidth -\horzsep,
legend style={legend columns=3,legend to name=legendname, legend cell align=left,/tikz/every even column/.append style={column sep=0.25cm}}
]
 \nextgroupplot[xmin=0.8,xmax=1.2,ymin=0.8,ymax=1.2,xtick=\empty,ytick=\empty,xlabel={$\nnInput_{(1)}$},ylabel={$\nnInput_{(2)}$},title={(a)}]
 \input{\basepath nnv-example_11.tikz}
\coordinate (top) at (rel axis cs:0,1);
\nextgroupplot[xmin=-0.35,xmax=-0.175,ymin=-1.75,ymax=-1.6,xtick=\empty,ytick=\empty,xlabel={$\nnHidden_{k(1)}$},ylabel={$\nnHidden_{k(2)}$},title={(b)}]
\input{\basepath nnv-example_16.tikz}
\nextgroupplot[xmin=2,xmax=2.55,ymin=1.35,ymax=1.42,xtick=\empty,ytick=\empty,xlabel={$\nnOutput_{(1)}$},ylabel={$\nnOutput_{(2)}$},title={(c)}]
\input{\basepath nnv-example_legends.tikz}
\input{\basepath nnv-example_19.tikz}
\coordinate (bot) at (rel axis cs:1,0);
\end{groupplot}
\path (top|-current bounding box.south)--coordinate(legendpos)(bot|-current bounding box.south);
\node at([yshift=-3ex]legendpos) {\pgfplotslegendfromname{legendname}};

\end{tikzpicture}%
%

    \caption{
        Verification of neural networks: (a) input to neural network, (b) output of hidden layer, and (c) output of neural network.
        As the enclosure does not intersect with the unsafe set, we can also conclude that the unobserved true output set does not intersect with the unsafe set~\eqref{eq:problem-statement}.
    }
    \label{fig:nnv-example}
\end{figure}

An example of this procedure is shown in \cref{fig:nnv-example}.
How tightly the obtained enclosures fit the true output sets heavily depends on the chosen set representation.
It has been shown that enclosures using (matrix) polynomial zonotopes work very well in practice as they obtain tight enclosures and are fast to compute~\cite{singh2018fast,ladner2023automatic,koller2025out}.
\section{Analog Neural Networks Under Varying Process Parameters}\label{sec:implementation-analog-nn}

\subsection{Schematic of the Analog Neuron Circuit}\label{sec:schematic-ann}
The schematic view of the analog neuron used in this work is illustrated in \cref{fig:overview},
which represents the main building block of our 130 nm, 1.2 V CMOS inference circuit \cite{aul2021schematic}.
The neuron circuit functionality can be broken down into three main parts:
(a) summation of input weighted currents with the bias current using Kirchhoff's laws,
(b) a ReLU activation function implemented using the left side of the current mirror to rectify negative currents,
and (c) the right side of the current mirror, which is utilized to scale the output current using a $5$-bit programmable weight $w_{(0{:}4)}$ and an additional bit $w_{(5)}$ to store the sign of the weight using another upper PMOS current mirror.
The neuron circuit accommodates an additional bias circuit that maps an $8$-bit programmable bias to an analog output current.
Neural network training is carried out off-chip using TensorFlow, followed by quantization to produce $6$-bit quantized weights in the range of $[-2,\,2]$.

Please note that in this circuit design, ReLU is applied before the weight multiplication.
This design is chosen to minimize energy usage.

However, neural networks typically start with a linear layer and do not apply ReLU on the input.
Thus, to maintain the inference circuit functionality for negative inputs,
the circuit schematic in \cref{fig:overview} is adapted to not apply a ReLU activation in the first layer and pass the negative input through instead.
Additionally, an output layer with a weight of $1$ is appended to apply the final ReLU activation at the network output if required by the nominal model.

\begin{figure}
    \centering
    \begin{circuitikz}[scale = 0.55, transform shape]  
\ctikzset{tripoles/mos style/arrows}

    \tikzset {
        activation/.style={circle,draw=black,inner sep=2px, minimum height=0.8cm, minimum width=0.8cm},
        relu/.pic = {
            \draw (-0.2,-0.1) -- (0,-0.1) -- (0.2,0.1);
        },
        weight/.style={circle,draw=black,inner sep=2px, minimum height=0.8cm, minimum width=0.8cm},
        weight/.pic = {
            \draw[ultra thin] (-0.2,-0.2) -- (0.2,-0.2);
            \draw[ultra thin] (-0.2,-0.2) -- (0.2,0);
            \draw[ultra thin] (-0.2,-0.2) -- (0.2,0.2);

            \draw[ultra thin] (-0.2,0) -- (0.2,-0.2);
            \draw[ultra thin] (-0.2,0) -- (0.2,0);
            \draw[ultra thin] (-0.2,0) -- (0.2,0.2);

            \draw[ultra thin] (-0.2,0.2) -- (0.2,-0.2);
            \draw[ultra thin] (-0.2,0.2) -- (0.2,0);
            \draw[ultra thin] (-0.2,0.2) -- (0.2,0.2);
        },
        bus/.style={line width=1.5pt}
    }

\tikzstyle{every node}=[font=\large]
% \fill [green!5] (-5.5,-3) rectangle (-4,8) node[midway,below=5.5cm,black,font=\small]{{(i) Sum}};
\draw  (-5.5,-3) rectangle (-3.8,8)[fill=CORAcolorBlue!20, line width=0.5pt, draw = none] node[midway,above=4cm,black,font=\large]{{(a)}};
% \draw (-4.75,6.75) node[activation] (sum) {\Large $\Sigma$};
% \fill[fill=yellow!8] (-3.8,-3) rectangle (-1.3,8) node[midway,below=5.5cm,black,font=\small]{{(ii) ReLU frontend}};
\draw  (-3.8,-3) rectangle (-1.3,8) [fill=CORAcolorBlue!20, line width=0.5pt, draw = none] node[midway,above=4cm,black,font=\large]{{(b)}};
% \draw (-2.55,6.75) node[activation] (relu){}; \path (relu.center) pic {relu};
% \fill [CORAcolorBlue!20] (-1.1,-3) rectangle (10.5,8) node[midway,below=5.5cm,black,font=\small]{{(iii) Weight cell}};
\draw  (-0.85,7.75) rectangle (10.75,-3.25) [fill=CORAcolorBlue!20, line width=0.5pt]; % node[anchor=south east,font=\tiny]{$W_{k(l,m)}$}
\draw  (-0.975,7.875) rectangle (10.625,-3.125) [fill=CORAcolorBlue!20,line width=0.5pt]; % node[anchor=south east,font=\tiny]{$W_{k(l,m)}$}
\draw  (-1.3,8) rectangle (10.5,-3) [fill=CORAcolorBlue!20, line width=0.5pt, draw = none] node[midway,above=4.325cm,left={4cm},black,font=\large]{{(c)}};
\draw  (-5.5,8) rectangle (10.5,-3) [line width=0.5pt] node[anchor=south east,font=\large]{$W_{k(i,j)}$};
\draw [dashed] (-3.8,-3) -- (-3.8,8);
\draw [dashed] (-1.2,-3) -- (-1.2,8);
% node[anchor=south east,font=\tiny]{$W_{k(i,j)}$};
% \draw (4.7,7) node[weight] (weight){}; \path (weight.center) pic {weight};
% \draw (4.7,6.75) node[weight,font=\large] (weight){$W_{k(i,j)}$};
% \draw  (-0.7,3.25) to[short] ++ (7.0,0) [line width=2.5pt];
% \draw  (-0.7,3.18) to[short] ++ (0.0,1.35) [line width=2.5pt];
% \draw  (-0.78,4.53) to[short] ++ (0.8,0.0) [line width=2.5pt];
\draw (0,4.5) -- (-0.7,4.5) -- (-0.7,3.29) -- (6.23,3.29) [bus];

\draw
(2.5,4.5)node[draw,minimum width=5.0cm,minimum height=1cm, ,font=\large] (reg) {6-bit register $w$}

%part 1
(-2.5,0)     node[nmos] (nmos02) {}  node[right] at (nmos02) {}
(nmos02.S) node[nmos, xscale=-1, anchor=D] (nmos01) {} node[] at (nmos01) {\small $1/1$}  %below left,yshift=0.3cm 
(nmos02.G) node[below,font=\large] {$V_{n}$}
(nmos01.S) node[sground] {}
(nmos02.D) node[pmos, anchor=D] (pmos03) {} node[right] at (pmos03) {}
(pmos03.G) node[below,font=\large] {$V_{p}$}
(nmos01.G) |- (pmos03.D) node[circ] {}
(pmos03.S) to[short] ++(0,0.5)
-- ++(-0.5,0) node[inputarrow,rotate=0] {} node [above,font=\large]{$I_\text{in}$} to[short] ++(-1.2,0) node[circ] {}
%node[inputarrow,rotate=-45] {} node [right,]{} to[short] ++(-1.0,1.0)
(-4.25,2.8) -- ++(-0.4,0) node[inputarrow,rotate=0] {} node [right,]{} to[short] ++(-0.6,0)
(-5.25,1.77) to[short] ++(0.6,0.6) node[inputarrow,rotate=45] {} node [right,]{} -- ++(0.4,0.4)
(-5.25,3.85) to[short] ++(0.6,-0.6) node[inputarrow,rotate=-45] {} node [right,]{}  -- ++(0.4,-0.4)

%part 2
(0.5,0)     node[nmos] (nmos2) {}  node[right] at (nmos2) {} 
(nmos2.S) node[nmos, xscale=1, anchor=D] (nmos1) {} node[] at (nmos1) {\small $1/1$}  %below left,yshift=0.3cm
($(nmos01)!0.5!(nmos1)$) ++(0.5,-1) node[font=\normalfont, fill=CORAcolorBlue!20, inner sep=0.1pt] {(NMOS current mirror)}
% (nmos1.S) node[sground] {}
(nmos2.D) node[nmos, anchor=S] (nmos3) {} node[right] at (nmos3) {}
(nmos3.G) to[short] node [right,font=\large]{$w_{(4)}$} ++(0,1.7)

(2,0)     node[nmos] (nmos22) {}  node[right] at (nmos22) {}
(nmos22.S) node[nmos, xscale=1, anchor=D] (nmos11) {} node[] at (nmos11) {\small $1/2$}  %below left,yshift=0.3cm
% (nmos11.S) node[sground] {}
(nmos11.S) -| (nmos1.S)
(nmos22.D) node[nmos, anchor=S] (nmos33) {} node[right] at (nmos33) {}
(nmos33.G) to[short] node [right,font=\large]{$w_{(3)}$} ++(0,1.7)

(nmos3.D) |- (nmos33.D) node[circ] {}
(nmos1.G)  ++ (0,0.75) to[short] ++(-0.55,0)
(nmos1.G)  ++ (0,0.75) to[short] ++(-1.05,0)  node[circ] {}
(nmos1.G) -| ++(0,0.75) node[circ] {} -| (nmos11.G)
(nmos2.G) -| ++(0,0.75) node[circ] {} -| (nmos22.G)
(nmos2.G)  ++ (0,0.75) to[short] ++(-0.25,0)
(nmos2.G)  ++ (0,0.75) to[short] ++(-0.25,0) node[below,font=\large] {$V_{n}$}

(3.5,0)     node[nmos] (nmos222) {}  node[right] at (nmos222) {}
(nmos222.S) node[nmos, xscale=1, anchor=D] (nmos111) {} node[] at (nmos111) {\small $1/4$}  %below left,yshift=0.3cm
(nmos111.S) node[sground] {}
(nmos111.S) -| (nmos11.S)
(nmos222.D) node[nmos, anchor=S] (nmos333) {} node[right] at (nmos333) {}
(nmos333.G) to[short] node [right,font=\large]{$w_{(2)}$}  ++(0,1.7)

(nmos33.D) |- (nmos333.D) node[circ] {}
(nmos11.G) -| ++(0,0.75) node[circ] {} -| (nmos111.G)
(nmos22.G) -| ++(0,0.75) node[circ] {} -| (nmos222.G)
%(nmos33.G) -| ++(0,1.75) node[circ] {} -| (nmos333.G)

(5,0)     node[nmos] (nmos2222) {}  node[right] at (nmos2222) {}
(nmos2222.S) node[nmos, xscale=1, anchor=D] (nmos1111) {} node[] at (nmos1111) {\small$1/8$}  %below left,yshift=0.3cm
% (nmos1111.S) node[sground] {}
(nmos1111.S) -| (nmos111.S)
(nmos2222.D) node[nmos, anchor=S] (nmos3333) {} node[right] at (nmos3333) {}
(nmos3333.G) to[short] node [right,font=\large]{$w_{(1)}$} ++(0,1.7)

(nmos333.D) |- (nmos3333.D) node[circ] {}
(nmos111.G) -| ++(0,0.75) node[circ] {} -| (nmos1111.G)
(nmos222.G) -| ++(0,0.75) node[circ] {} -| (nmos2222.G)
%(nmos333.G) -| ++(0,1.75) node[circ] {} -| (nmos3333.G)

(6.5,0)     node[nmos] (nmos22222) {}  node[right] at (nmos22222) {} 
(nmos22222.S) node[nmos, xscale=1, anchor=D] (nmos11111) {} node[] at (nmos11111) {\small $1/16$} node[xshift=2cm] at (nmos11111) {\normalfont (W/L relations)}  % TL: made it \small as it was too prominent otherwise;
% (nmos11111.S) node[sground] {}
(nmos1111.S) -| (nmos11111.S)
(nmos22222.D) node[nmos, anchor=S] (nmos33333) {} node[right] at (nmos33333) {}
(nmos33333.G) to[short] node [right,font=\large]{$w_{(0)}$} ++(0,1.7)

(nmos3333.D) |- (nmos33333.D) node[circ] {}
(nmos1111.G) -| ++(0,0.75) node[circ] {} -| (nmos11111.G)
(nmos2222.G) -| ++(0,0.75) node[circ] {} -| (nmos22222.G)
%(nmos3333.G) -| ++(0,1.75) node[circ] {} -| (nmos33333.G)

%part 3
(9,4)     node[pmos] (pmos2) {}  node[right] at (pmos2) {}
(pmos2.G) node[pmos, xscale=-1, anchor=G] (pmos1) {} node[left] at (pmos1){}
(pmos2.S) node[pmos, anchor=D] (pmos3) {} node[right] at (pmos3) {}
(pmos3.G) node[pmos, xscale=-1, anchor=G] (pmos4) {} node[left] at (pmos4){}
(pmos3.S) node[pmos, anchor=D] (pmos5) {} node[right] at (pmos5) {}
(pmos5.G) node[pmos, xscale=-1, anchor=G] (pmos6) {} node[left] at (pmos6){}
(pmos5.S) -| (pmos6.S)
(pmos6.S) node[anchor=north east,font=\large] {$V_\text{dd}$}
(8.03,2.31) node[nmos, yscale=-1, rotate=90] (nmosx) {} node[left] at (nmosx){}
(nmos33333.D) -| ++(0.55,0) node[circ] {} (nmosx.D)
%(nmos33333.G)  -| ++(0,1.73) node[circ] {} |- (nmosx.G)
(nmosx.S) -| (pmos2.D)
(pmos1.D) |- (nmosx.D)
(pmos1.G) node[circ] {} -| (nmosx.G) node[circ] {}
(7.9,5.55) node[circ] {} |- ++(-1.8,0.75) node[below,font=\large] {$V_\text{p2}$}
($(pmos6.S)!0.65!(6.1,6.3)$) ++(-0,0) node[anchor=east,font=\normalfont] {(PMOS current mirror)}
(nmosx.G) to[short] node [above,font=\large]{$w_{(5)}$} ++(-2,0)
(pmos5.G) node[circ] {} |- (pmos4.D) node[circ] {}
(nmosx.S) ++(0.2,0) node[circ] {} to[short] ++(1.0,0)
node[inputarrow,rotate=0] {} node [below,font=\large]{$I_\text{out}$} to[short] ++(0,0)
    ;
\end{circuitikz}
    \caption{Schematic of the analog neuron circuit:
            (a) current summation block,
            (b) ReLU activation function, and
            (c) a cell for each weight $W_{k(i,j)}$.
            The programmable weight bits $w_{(0{:}5)}$ are stored in a $6$-bit register.
            The bit~$w_{(5)}$ is used to realize negative weights by inverting the output current.
    }
    \label{fig:overview}
\end{figure}

\subsection{Performance Analysis}

\begin{figure}[t]
\centering
    \tikzsetnextfilename{./figures/Process_params1/process_params1}%
    \definecolor{mycolor1}{rgb}{0.00000,0.44700,0.74100}%
\begin{tikzpicture}
\footnotesize
\pgfplotsset{
plotstyle1/.style={mark=no, draw=none, fill=CORAcolorBlue, forget plot},
plotstyle2/.style={only marks, mark=*, mark options={}, mark size=0.1000pt, legend image post style={mark size=1pt}, draw=black, forget plot}
}
\def\rows{1}
\def\cols{2}
\def\horzsep{1.5cm}
\def\basepath{./figures/Process_params1/}

\begin{groupplot}[%
group style={rows = \rows, columns = \cols, horizontal sep = \horzsep},
scale only axis,
width=1/\cols*\linewidth -\horzsep,
legend style={legend columns=2,legend to name=legendname, legend cell align=left,/tikz/every even column/.append style={column sep=0.5cm}}
]
\nextgroupplot[xmin=50,xmax=75,ymin=0,ymax=180,xlabel={Output current [nA]},ylabel={Nr.\ of samples},title={(a)}]
\input{\basepath process_params1_11.tikz}
\coordinate (top) at (rel axis cs:0,1);
\nextgroupplot[xmin=-200,xmax=400,ymin=0,ymax=450,xlabel={Input current [nA]},ylabel={Output current [nA]},title={(b)}]
\input{\basepath process_params1_legends.tikz}
\input{\basepath process_params1_12.tikz}
\coordinate (bot) at (rel axis cs:1,0);
\end{groupplot}
\path (top|-current bounding box.south)--coordinate(legendpos)(bot|-current bounding box.south);
% \node at([yshift=-6ex]legendpos) {\pgfplotslegendfromname{legendname}};

\end{tikzpicture}%
%

\caption{Monte Carlo analysis of the neuron netlist under process
variations: (a) parameter variations in a single neuron for an input current of $50$ [nA], (b) output of the neuron impacted by all process parameters.}
    \label{fig:Process_params1}
\end{figure}

Let us first analyze the performance of the circuit under varying process parameters to establish an accurate model of the circuit.
In \cref{fig:Process_params1}, we visualize Monte Carlo simulations of the neuron netlist under process variations.
\Cref{fig:Process_params1}a shows an irregular variation distribution contrary to the normal distribution observed for device mismatch~\cite[Fig.~4]{ladner2025analog}. 
We also observe that the distribution varies for each of the $64$ quantized weight values.
This is expected because different bits of the register activate different transistor groups.
Additionally, the circuit behavior deviates depending on the sign of the input to the neuron.
All the aforementioned challenging aspects lead to the need for deriving a unique parametric model for each of the $64$ weights. 

\subsection{Sensitivity Modeling}
To identify the impact of varying process parameters on our analog neuron circuit,
we run a sensitivity analysis with an underlying real process based on the PSP transistor model.
As shown in \cref{fig:overview}, the neuron circuit contains both NMOS and PMOS transistors operating in the subthreshold region, 
which are subject to more than $50$ model parameters.

Our sensitivity analysis runs Monte Carlo sampling with process variations on the neuron netlist and analyzes the variance contributions to the output current from all process parameters.
Afterward, we choose the parameters with the highest contribution as the dominant parameters.
This step has to be done only once per neuron netlist and is reused across all network architectures built from it.
For our neuron netlist (\cref{fig:overview}), the behavior can be well described using two dominant parameters, $\phi_1$ and $\phi_2$,
which represent two parameters influencing the surface potential of the NMOS transistor, affecting the threshold voltage and the overall transistor behavior.
The strong influence of these parameters on the output current is shown in \cref{fig:Process_params2}.
While $\phi_1,\,\phi_2$ greatly influence NMOS transistors, they only have a negligible effect on PMOS transistors since the current mirror topology mitigates the impact of surface potential variations.

\begin{figure}[t]
\centering
    \tikzsetnextfilename{./figures/Process_params2/process_params2_new}%
    \definecolor{mycolor1}{rgb}{0.00000,0.44700,0.74100}%
\begin{tikzpicture}
\footnotesize
\pgfplotsset{
    plotstyle1/.style={only marks, mark=*, mark options={}, mark size=0.1000pt, legend image post style={mark size=1pt}, draw=black, forget plot}
}
\def\rows{1}
\def\cols{2}
\def\horzsep{1.5cm}
\def\basepath{./figures/process_params2/}

\begin{groupplot}[%
group style={rows = \rows, columns = \cols, horizontal sep = \horzsep},
scale only axis,
width=1/\cols*\linewidth -\horzsep,
legend style={legend columns=2,legend to name=legendname, legend cell align=left,/tikz/every even column/.append style={column sep=0.5cm}}
]
\nextgroupplot[xmin=0.6,xmax=1.3,ymin=52,ymax=72,xlabel={$\phi_1$},ylabel={Output current [nA]}]
\input{\basepath process_params2_new_11.tikz}
\coordinate (top) at (rel axis cs:0,1);
\nextgroupplot[xmin=0.8,xmax=1.2,ymin=52,ymax=72,xlabel={$\phi_2$},ylabel={Output current [nA]}]
\input{\basepath process_params2_new_legends.tikz}
\input{\basepath process_params2_new_12.tikz}
\coordinate (bot) at (rel axis cs:1,0);
\end{groupplot}
\path (top|-current bounding box.south)--coordinate(legendpos)(bot|-current bounding box.south);
% \node at([yshift=-6ex]legendpos) {\pgfplotslegendfromname{legendname}};

\end{tikzpicture}%
%

\caption{
    Monte Carlo analysis of the output current of the neuron netlist versus the two dominant parameters:
    (a) $\phi_1$ parameter variations, and (b) $\phi_2$ parameter variations,
    where $\phi_1,\phi_2$ are sampled from a normal distribution.
    }
    \label{fig:Process_params2}
\end{figure}

\section{Bounding Process Variations Using Set-Based Computing}

In this section, we first establish a model describing the effect of process variations on the output of an analog neural network, followed by a description of how these variations can be bounded using set-based computing.

\subsection{Modeling Process Variations}
We model the uncertainty of the dominant parameters $\phi_1$ and $\phi_2$ around their nominal value of 1:
\begin{align}
    \label{eq:process-model-phis}
    \begin{split}
        \phi_1 &\in\mathcal{D}_1\subset\R, \\
        \phi_2 &\in\mathcal{D}_2\subset\R,
    \end{split}
\end{align}
where the domains $\mathcal{D}_1,\mathcal{D}_2$ are chosen to enclose a desired number of Monte Carlo simulations (\cref{fig:Process_params2}).
In particular, the process parameters $\phi_1,\phi_2$ are sampled from a normal distribution with standard deviations $\sigma_1$ and $\sigma_2$, respectively,
and we choose, e.g., $\mathcal{D}_1=[1-3\sigma_1,1+3\sigma_1]$ to enclose $\approx99\%$ of the simulations.
The subtle variation produced by the rest of the varying process parameters is captured with a single third noise symbol $\phi_3\in\mathcal{D}_3\subset\R$.
By choosing the domains large enough such that all measurements are contained, we say that the model is reachset conformant~\cite{roehm2019model}.

With these variations modeled, we can describe a neural network with respect to $\phi_1$, $\phi_2$ and account for all error terms.
To distinguish between the analog implementation and the nominal neural network (\cref{def:neural-networks}),
we decorate the respective variables of the analog implementation with a tilde: $\widetilde{h}_k\in\R^{\numNeurons_k}$, $\widetilde{W}_k\in\R^{\numNeurons_{k}\times\numNeurons_{k-1}}$, $\widetilde{b}_k\in\R^{\numNeurons_k}$, $k\in[\numLayers]$.
The weights $\widetilde{W}_k$ are influenced polynomially by $\phi_1$ and $\phi_2$,
and we obtain the respective coefficients $a_{k,l,m},\,b_{k,l,m}\in\R$ of each of the $64$ quantized weights of the nominal model individually using polynomial regression.
Each entry $\widetilde{w}_k$ of $\widetilde{W}_k$ is modeled as:
\begin{align}
    \label{eq:process-model-weights}
    \widetilde{w}_{k} &= \sum_{l=0}^{3} \sum_{m=0}^{3} a_{k,l,m} {\phi_1}^{l} {\phi_2}^{m} + \phi_3, \quad l+m \leq 3.
\end{align}
The influence of $\phi_1,\phi_2$ on the bias $\widetilde{b}_k$ is negligible according to the polynomial regression.
Moreover, for negative inputs to ReLU activations, the output of the respective neuron is set to:
\begin{align}
    \label{eq:process-model-polynomial}
    \widetilde{\nnHidden}_{k} = \sum_{l=0}^{3} \sum_{m=0}^{3} b_{k,l,m} {\phi_1}^{l} {\phi_2}^{m}, \quad l+m \leq 3,
\end{align}
to account for the rectified output.
As described in \cref{sec:schematic-ann}, we deploy a slightly different circuit for the first layer to deal with negative inputs;
thus, the coefficients for the respective weights also differ slightly from the ones in all other layers.
The polynomial model is both simple and scalable, as determining the coefficients for each weight only needs to be done once and can be reused for all architectures implemented using the same netlist (\cref{fig:overview}).

We also analyze the influence of process variations on convolutional neural networks~\cite[Sec.~5.5.6]{bishop2006pattern}, 
and we find that our model also generalizes to convolutional neural networks and only requires updating the coefficients $a_{k,l,m},\,b_{k,l,m}$.
Please note that, as a convolution is applied---which has far fewer weights than a matrix multiplication---%
the overall variation is typically much smaller for convolutional neural networks.
Thus, convolutional neural networks are more robust against process variations.

% CONVOLUTIONAL
        % \widetilde{\nnHidden_k} &= \ReLU(\widetilde{W_k} \widetilde{\nnHidden_{k-1}} + b_k),\\
 %       \text{Conv. layer: }
 %       \widetilde{\nnHidden}_1 &= \begin{cases}
 %  \widetilde{\nnHidden}_{0} \circledast \widetilde{W}'_1 + b_k & \text{if $x < 0$} \\
 %    \widetilde{\nnHidden}_{0} \circledast \widetilde{W}_1 + b_k & \text{otherwise}
 % \end{cases}\\
 %       \widetilde{\nnHidden}_k &= \widetilde{\nnHidden}_{k-1} \circledast \widetilde{W}_k + b_k,\\

% -------------------------------------------------------------------

\subsection{Bounding Process Variations}
\label{sec:bounding-process-variations}
Process variations can be bounded by following the general verification procedure described in \cref{sec:verification-neural-network}.
We start the verification by propagating the input set $\widetilde{\nnHiddenSet}_0=\nnInputSet$ through the first layer.
In contrast to the general procedure,
the weights $\widetilde{W}_1$ of linear layers~\eqref{eq:process-model-weights} are no longer matrices but sets $\widetilde{\mathcal{W}}_1$ due to the uncertainties introduced in \eqref{eq:process-model-phis}.
Our polynomial model~\eqref{eq:process-model-weights} enables us to represent $\widetilde{\mathcal{W}}_1$ as polynomial zonotopes,
where the noise symbols $\phi_1,\,\phi_2,\,\phi_3$ become (scaled) dependent factors $\alpha_1,\,\alpha_2,\,\alpha_3$, as introduced in \cref{def:pZ-mat}.

\begin{figure}[t]
\centering
    \tikzsetnextfilename{./figures/poly-enclosure}%
    \input{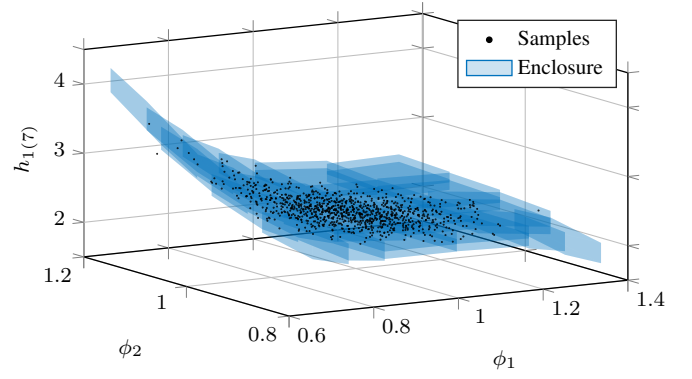}%

\caption{
    Polynomial zonotope enclosure of neuron $7$ of the first hidden layer---having the maximal value according to the nominal model---with respect to the main process parameters $\phi_1,\phi_2$.
}
    \label{fig:poly-enclosure}
\end{figure}

Given $\widetilde{\nnHiddenSet}_0$ and $\widetilde{\mathcal{W}}_1$, their multiplication is computed using~\eqref{eq:pz-multiplication}.
An example enclosure of the first linear layer is visualized in \cref{fig:poly-enclosure},
where the enclosure is shown with respect to the main process parameters $\phi_1,\phi_2$.
As these two parameters describe the behavior well, i.e., low variance for each ($\phi_1,\phi_2$)-pair, 
the enclosure barely has to be enlarged in the z-direction to enclose the sampled simulations.
This confirms our analysis that the remaining process variations barely influence the final outcome.
Please note that the enclosure is represented using a single set $\widetilde{\nnHiddenSet}_1$ and is only split for visualization purposes.

As we proceed to propagate the set through the entire network,
we repeatedly multiply $\widetilde{\mathcal{W}}_k$ with the current set $\widetilde{\nnHiddenSet}_{k-1}$, $k\in[\numLayers]$,
and enclose the ReLU nonlinearities using standard techniques (\cref{prop:nnv-enclosure}),
until we obtain an enclosure of the output set $\widetilde{\nnOutputSet}$ to verify given specifications~\eqref{eq:problem-statement}.

\section{Experimental Results}\label{sec:experiments}

To evaluate our modeling approach, we use three different classification datasets: Breast Cancer \cite{wisconsin}, Iris \cite{iris_53}, and MNIST \cite{lecun2010mnist}. We simulate our analog inference circuit and run Monte Carlo simulations in Cadence Spectre~\cite{Cadence24} on a workstation with dual Intel Xeon E5-2683 v4 CPUs ($16$ cores each). We run the set-based modeling approach implemented in MATLAB using the CORA toolbox \cite{althoff2015introduction, Althoff2025manual} on a home computer with an Intel Core i7-1165G7 CPU.
To evaluate the effectiveness of our modeling approach for each dataset, we compare it with $1,000$ Monte Carlo simulations with inputs taken from the test set.
If not stated otherwise, all figures and tables are with respect to a $3\sigma$ confidence set~\eqref{eq:process-model-phis}.

\subsection{Main Results}

We report the required time to analyze the circuit under process variations in \cref{tab:mnist-iris-comparison}.
While the Monte Carlo simulations require hours for the Breast Cancer examples and even up to a day for the MNIST CNN example, our approach runs in a few seconds.
It is worth mentioning that we require a setup time of $12$ minutes to extract the parametric model~\eqref{eq:process-model-weights} for the netlist (\cref{fig:overview})
by running Monte Carlo simulations on a single neuron for $64$ different weight values for positive and then negative input individually.
However, this is only required once and can be reused for all architectures.

As shown in \cref{tab:mnist-iris-comparison}, we are able to enclose on average around $99 \%$ of the output variation samples,
as expected from the enclosed number of Monte Carlo samples using a $3\sigma$ confidence set~\eqref{eq:process-model-phis}.
While the considered models are relatively small, we want to stress that this is not a limitation of our verification approach~\cite{kaulen20256th}.
Formal verification scales to much larger networks~\cite{kaulen20256th};
however, the perturbations occurring in the analog implementation become too large to make useful predictions.
For example, the Breast Cancer neural network only consists of three layers, and we have already observed large perturbations on the output for such sizes.

However, our verification approach can drastically reduce the design cycle of these larger models by avoiding the time-consuming Monte Carlo simulations.
Using \cref{tab:verification-comparison}, we can quickly assess the expected performance of the manufactured circuits.
For example, circuits deviating up to $2\sigma$ retain a verified accuracy competitive with the accuracy of the nominal model,
meaning that $\approx 95\%$ of the circuits can be used in production and only $\approx 5\%$ may be faulty.
Moreover, the results highlight the need for a more robust circuit design to obtain fewer faults ($3\sigma$),
and any new circuit design can quickly be validated as demonstrated in \cref{tab:mnist-iris-comparison}.

\begin{table}
    \caption{Time comparison and average enclosure of Monte Carlo simulations of our approach per input pattern.}
    \centering
    \newcommand{\inv}[1]{{\hphantom{#1}}}
    \begin{tabular}{c c l r c }
        \toprule
        \textbf{Dataset}             & \textbf{\#Classes}  & \textbf{Approach} & \textbf{Time [s]} & \textbf{Enclosure} \\
        \midrule
        \multirow{2}*{Breast Cancer} & \multirow{2}*{$2$}  & Cadence           & $5,479.72$        & --                 \\
        &                     & Ours              & $9.50$            & $99.08 \%$         \\
        \midrule
        \multirow{2}*{Iris}          & \multirow{2}*{$3$}  & Cadence           & $865.64$          & --                 \\
        &                     & Ours              & $1.15$            & $98.01 \%$         \\
        \midrule
        \multirow{2}*{MNIST}         & \multirow{2}*{$10$} & Cadence           & $29,592.60$       & --                 \\
        &                     & Ours              & $1.77$            & $97.77 \%$         \\
        \midrule
        MNIST                        & \multirow{2}*{$10$} & Cadence           & $84,409.00$       & --                 \\
        CNN                          &                     & Ours              & $3.37$            & $99.51 \%$         \\
        \bottomrule
    \end{tabular}
    \label{tab:mnist-iris-comparison}
\end{table}

\begin{table}
    \caption{
        Nominal and verified accuracy under process variations on the input patterns of all test datasets with varying desired enclosure of the Monte Carlo simulations via the standard deviations $\sigma$~\eqref{eq:process-model-phis}.
    }
    \centering
    \newcommand{\inv}[1]{{\hphantom{#1}}}
    \begin{tabular}{c c c c c c }
        \toprule
        \textbf{Dataset} & \textbf{\#Patterns} & \textbf{Nominal} & $\mathbf{1 \sigma}$ & $\mathbf{2 \sigma}$& $\mathbf{3 \sigma}$  \\
        \midrule
        Breast Cancer    & $100$               & $98\%$           & $96\%$              & $95\%$              & $27\%$              \\
        % \midrule
        Iris             & $75$                & $92\%$           & $88 \%$             & $83 \%$             & $48 \%$             \\
        % \midrule
        MNIST            & $100$               & $87\%$           & $78 \%$             & $71 \%$             & $59 \%$             \\
        % \midrule
        MNIST CNN        & $100$               & $93\% $          & $81\% $             & $64 \%$             & $28\% $             \\
%          \midrule
%          Avg.  &        & $93\%$ & $86\%$ & $78\%$ & $41\%$ \\
        \bottomrule
    \end{tabular}
    \label{tab:verification-comparison}
\end{table}

\subsection{Additional Experiments and Ablation Studies}

\subsubsection{Computing Enclosures and Verified Accuracies}

In this experiment, we detail the training process and results for each dataset.

\begin{figure}[t]
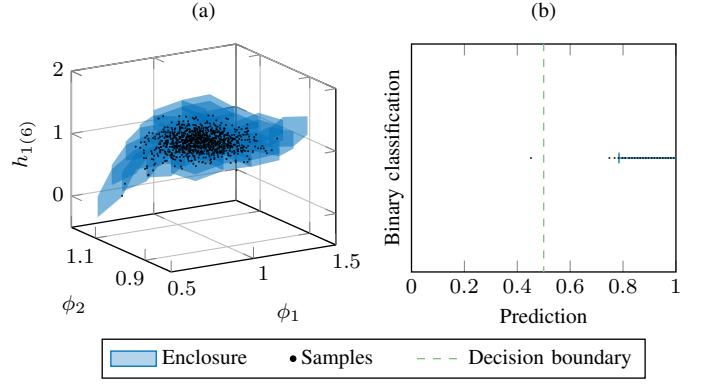

    \centering
    \tikzsetnextfilename{./figures/Results/breastcancer_results}%
    \definecolor{mycolor1}{rgb}{0.85000,0.32500,0.09800}%
\begin{tikzpicture}
\footnotesize
\pgfplotsset{
plotstyle1/.style={color=black, only marks, mark=*,mark size=0.1000pt, legend image post style={mark size=1pt}, mark options={solid, black}},
plotstyle2/.style={area legend, draw=none, fill=CORAcolorBlue, fill opacity=0.3, legend image post style={draw=CORAcolorBlue}, forget plot},
plotstyle3/.style={color=CORAcolorBlue, mark=|, legend image post style={mark=8}, forget plot},
plotstyle4/.style={only marks, mark=*, mark options={}, mark size=0.1000pt, legend image post style={mark size=1pt}, draw=black, forget plot},
plotstyle5/.style={draw=CORAcolorSafe,dashed,forget plot}
}
\def\rows{1}
\def\cols{2}
\def\horzsep{1cm}
\def\basepath{./figures/Results/}

\begin{groupplot}[%
group style={rows = \rows, columns = \cols, horizontal sep = \horzsep},
scale only axis,
width=1/\cols*\linewidth -\horzsep,
legend style={legend columns=3,legend to name=legendname, legend cell align=left,/tikz/every even column/.append style={column sep=0.5cm}}
]
\nextgroupplot[xmin=0.5,xmax=1.5,ytick={0.0.75,1.0,1.25},ymin=0.794883548,ymax=1.1968,ytick={0.9,1.1},zmin=-0.5,zmax=2,xlabel={$\phi_1$},ylabel={$\phi_2$},zlabel={$\nnHidden_{1(6)}$},title={(a)},view={-31.2123636430442}{18.4665500565668},xmajorgrids,ymajorgrids,zmajorgrids]
\input{\basepath breastcancer_results_11.tikz}
\coordinate (top) at (rel axis cs:0,1);
\nextgroupplot[xmin=0,xmax=1,ymin=0,ymax=2,ytick=\empty,xlabel={Prediction},ylabel={Binary classification},title={(b)}]
\input{\basepath breastcancer_results_legends.tikz}
\input{\basepath breastcancer_results_12.tikz}
\coordinate (bot) at (rel axis cs:1,0);
\end{groupplot}
\path (top|-current bounding box.south)--coordinate(legendpos)(bot|-current bounding box.south);
\node at([yshift=-3ex]legendpos) {\pgfplotslegendfromname{legendname}};

\end{tikzpicture}%
%

    \caption{Enclosure of breast cancer dataset: (a) enclosure of neuron $6$ of the first hidden layer---having the maximal value according to the nominal model---with respect to the two dominant process parameters, and (b) enclosure of the output layer Monte Carlo samples.}
    \label{fig:BR_results}
\end{figure}

\paragraph{Diagnostic Breast Cancer Dataset}
We trained the Diagnostic Breast Cancer Dataset in TensorFlow using a three-layer feedforward neural network of $30$ inputs, a fully-connected layer with $8$ hidden neurons, followed by another fully-connected layer with $16$ hidden neurons, and a single output neuron for binary classification.
An example enclosure with respect to the two dominant parameters $\phi_1$ and $\phi_2$ after the first hidden layer is depicted in \cref{fig:BR_results}a,
and the enclosure of the entire network is depicted in \cref{fig:BR_results}b.
We enclose $99.8 \%$ of the Monte Carlo samples in the output space and 
successfully verify the output simulations with label ``1'' as the lower bound of our obtained enclosure is above the classification threshold of $0.5$.
Doing so for all input patterns yields the verified accuracy.

\begin{figure}[t]
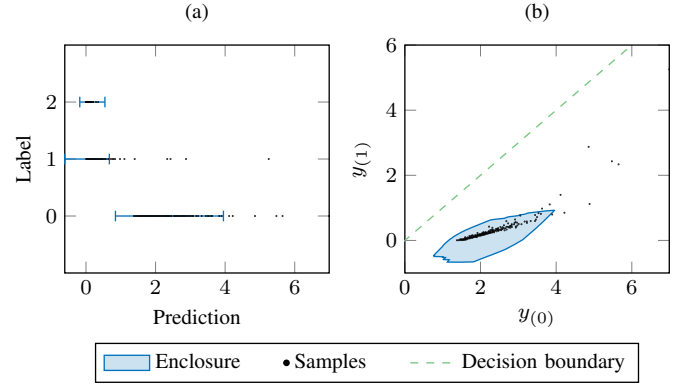

    \centering
    \tikzsetnextfilename{./figures/Results/Iris_results}%
    \begin{tikzpicture}
\footnotesize
\pgfplotsset{
plotstyle1/.style={color=CORAcolorBlue, mark=|, forget plot},
plotstyle2/.style={only marks, mark=*, mark options={},mark size=0.1000pt, legend image post style={mark size=1pt}, draw=black, forget plot},
plotstyle3/.style={area legend, draw=CORAcolorBlue, fill=CORAcolorBlue, fill opacity=0.2},
plotstyle4/.style={only marks, mark=*, mark options={}, mark size=0.1000pt, legend image post style={mark size=1pt}, draw=black},
 plotstyle5/.style={draw=CORAcolorSafe,dashed,forget plot}
}
\def\rows{1}
\def\cols{2}
\def\horzsep{1cm}
\def\basepath{./figures/Results/}

\begin{groupplot}[%
group style={rows = \rows, columns = \cols, horizontal sep = \horzsep},
scale only axis,
width=1/\cols*\linewidth -\horzsep,
legend style={legend columns=3,legend to name=legendname, legend cell align=left,/tikz/every even column/.append style={column sep=0.5cm}}
]
\nextgroupplot[xmin=-0.609857803005858,xmax=6.99104919727262,ymin=-1,ymax=3,ytick={0,1,2},xlabel={Prediction},ylabel={Label},title={(a)}]
\input{\basepath Iris_results_11.tikz}
\coordinate (top) at (rel axis cs:0,1);
\nextgroupplot[xmin=0,xmax=6.99104919727262,ymin=-1,ymax=6,xlabel={$y_{(0)}$},ylabel={$y_{(1)}$},title={(b)}]
\input{\basepath Iris_results_legends.tikz}
 \input{\basepath Iris_results_12.tikz}
\coordinate (bot) at (rel axis cs:1,0);
\end{groupplot}
\path (top|-current bounding box.south)--coordinate(legendpos)(bot|-current bounding box.south);
\node at([yshift=-3ex]legendpos) {\pgfplotslegendfromname{legendname}};

\end{tikzpicture}%
%

    \caption{Enclosure of Iris dataset: (a) projection of all dimensions, and (b) enclosure of the two most significant dimensions.}
    \label{fig:Iris_results}
\end{figure}

\begin{figure}[t]
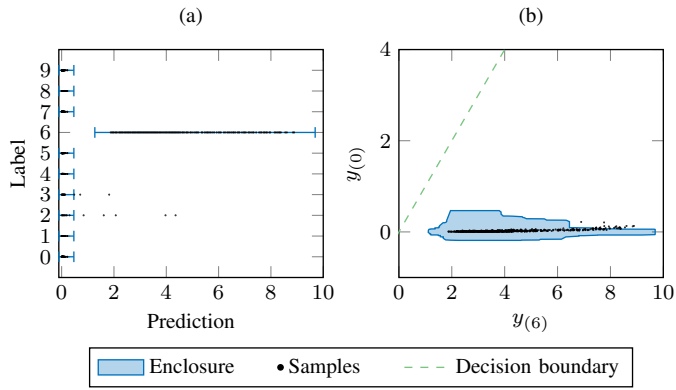

    \centering
    \tikzsetnextfilename{./figures/Results/mnist_dense_results}%
    \definecolor{mycolor1}{rgb}{0.85000,0.32500,0.09800}%
\definecolor{mycolor2}{rgb}{0.00000,0.44700,0.74100}%
\begin{tikzpicture}
\footnotesize
\pgfplotsset{
    plotstyle2/.style={color=black, only marks, mark=*,mark size=0.1000pt, legend image post style={mark size=1pt}, mark options={solid, black}},
    plotstyle3/.style={area legend, draw=CORAcolorBlue, fill=CORAcolorBlue, fill opacity=0.3, forget plot},
    plotstyle1/.style={color=CORAcolorBlue, mark=|, legend image post style={mark=8}, forget plot},
    plotstyle4/.style={only marks, mark=*, mark options={}, mark size=0.1000pt, legend image post style={mark size=1pt}, draw=black, forget plot},
    plotstyle5/.style={draw=CORAcolorSafe,dashed,forget plot}
}
\def\rows{1}
\def\cols{2}
\def\horzsep{1cm}
\def\basepath{./figures/Results/}

\begin{groupplot}[%
group style={rows = \rows, columns = \cols, horizontal sep = \horzsep},
scale only axis,
width=1/\cols*\linewidth -\horzsep,
legend style={legend columns=3,legend to name=legendname, legend cell align=left,/tikz/every even column/.append style={column sep=0.5cm}}
]
\nextgroupplot[xmin=-0.108036064899897,xmax=10,ytick={0,1,2,3,4,5,6,7,8,9},ymin=-1,ymax=10,xlabel={Prediction},ylabel={Label},title={(a)}]
\input{\basepath mnist_dense_results_11.tikz}
\coordinate (top) at (rel axis cs:0,1);
\nextgroupplot[xmin=0,xmax=10,ymin=-1,ymax=4,xlabel={$y_{(6)}$},ylabel={$y_{(0)}$},title={(b)}]
\input{\basepath mnist_dense_results_legends.tikz}
\input{\basepath mnist_dense_results_12.tikz}
\coordinate (bot) at (rel axis cs:1,0);
\end{groupplot}
\path (top|-current bounding box.south)--coordinate(legendpos)(bot|-current bounding box.south);
\node at([yshift=-3ex]legendpos) {\pgfplotslegendfromname{legendname}};

\end{tikzpicture}%
%

    \caption{Enclosure of the fully-connected network on MNIST: (a) projection of all dimensions, and (b) enclosure of the two most significant dimensions.}
    \label{fig:mnist_dense_results}
\end{figure}

\begin{figure}[t]
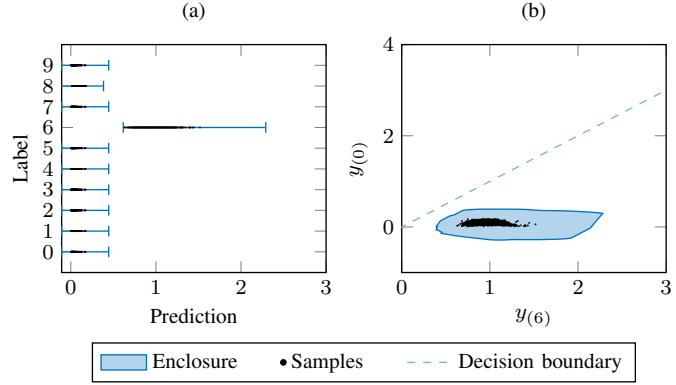

    \centering
    \tikzsetnextfilename{./figures/Results/mnist_cnn_results}%
    \definecolor{mycolor1}{rgb}{0.85000,0.32500,0.09800}%
\definecolor{mycolor2}{rgb}{0.00000,0.44700,0.74100}%
\begin{tikzpicture}
    \footnotesize
    \pgfplotsset{
        plotstyle2/.style={color=black, only marks, mark=*,mark size=0.1000pt, legend image post style={mark size=1pt}, mark options={solid, black}},
        plotstyle3/.style={area legend, draw=CORAcolorBlue, fill=CORAcolorBlue, fill opacity=0.3, forget plot},
        plotstyle1/.style={color=CORAcolorBlue, mark=|, legend image post style={mark=8}, forget plot},
        plotstyle4/.style={only marks, mark=*, mark options={}, mark size=0.1000pt, legend image post style={mark size=1pt}, draw=black, forget plot},
        plotstyle5/.style={draw=CORAcolorSafe,dashed,forget plot}
    }
    \def\rows{1}
    \def\cols{2}
    \def\horzsep{1cm}
    \def\basepath{./figures/Results/}

    \begin{groupplot}[%
        group style={rows = \rows, columns = \cols, horizontal sep = \horzsep},
        scale only axis,
        width=1/\cols*\linewidth -\horzsep,
        legend style={legend columns=3,legend to name=legendname, legend cell align=left,/tikz/every even column/.append style={column sep=0.5cm}}
    ]
        \nextgroupplot[xmin=-0.108036064899897,xmax=3,ytick={0,1,2,3,4,5,6,7,8,9},ymin=-1,ymax=10,xlabel={Prediction},ylabel={Label},title={(a)}]
        \input{\basepath mnist_cnn_results_11.tikz}
        \coordinate (top) at (rel axis cs:0,1);
        \nextgroupplot[xmin=0,xmax=3,ymin=-1,ymax=4,xlabel={$y_{(6)}$},ylabel={$y_{(0)}$},title={(b)}]
        \input{\basepath mnist_cnn_results_legends.tikz}
        \input{\basepath mnist_cnn_results_12.tikz}
        \coordinate (bot) at (rel axis cs:1,0);
    \end{groupplot}
    \path (top|-current bounding box.south)--coordinate(legendpos)(bot|-current bounding box.south);
    \node at([yshift=-3ex]legendpos) {\pgfplotslegendfromname{legendname}};

\end{tikzpicture}%
%

    \caption{Enclosure of the convolutional network on MNIST: (a) projection of all dimensions, and (b) enclosure of the two most significant dimensions.}
    \label{fig:mnist_cnn_results}
\end{figure}

\paragraph{Iris Dataset}
The Iris classification network has $4$ inputs, a fully-connected layer with $8$ hidden neurons, and an output layer with $3$ neurons representing the three classes. 
\cref{fig:Iris_results}a shows the ranges of the individual outputs representing each class. 
The bounds of the computed output enclosure are shown as intervals for each class. 
As the bounds for the correct label ``0'' do not overlap with the other classes, we have verified the example~\cite[Prop.~B.2]{ladner2023automatic}.
Additionally, in \cref{fig:Iris_results}b, we showcase the enclosure of the two largest dimensions,
where we enclose $99\%$ of the samples, again showing that our enclosure does not cross the decision boundary between the two classes.

% must stay in the first column of the last content page for balancing to apply there
\balance
\paragraph{MNIST Dataset}
For the MNIST handwritten digit recognition benchmark, we implemented two different neural network architectures: a fully-connected network and a convolutional neural network.
We reduced the network input size to $14\times14$ to decrease the circuit complexity and energy consumption.
The fully-connected network consists of a fully-connected layer with $10$ hidden neurons followed by an output layer with $10$ neurons representing the digit classes.
Furthermore, the convolutional network consists of a convolutional layer with a single filter of size $7\times7$, followed by a fully-connected layer of $64$ neurons, and finally an output layer with $10$ outputs.
As shown in \cref{fig:mnist_dense_results}a, the output of the fully-connected network shows a high sensitivity to process variations.
Nevertheless, the polynomial zonotope is able to verify the specification while enclosing $99.1 \%$ of the samples.
The convolutional neural network in \cref{fig:mnist_cnn_results}a exhibits much less sensitivity (note scaling of x-axis), where our model again verifies the specification and successfully encloses all the samples.
In \cref{fig:mnist_dense_results}b and \cref{fig:mnist_cnn_results}b, we again confirm the aforementioned results by showing the enclosure of the most significant outputs for each architecture.

\subsubsection{Comparison to Non-Polynomial Modeling}

Finally, we demonstrate that it is indeed necessary to have a polynomial model and that just using linear relaxations is not sufficient.
To do so, we adapt the previous approach on mismatch variations~\cite{ladner2025analog} based on zonotopes~\cite{girard2005reachability} and apply it to process variations.
In contrast to polynomial zonotopes (\cref{def:pZ-mat}), zonotopes can only represent convex sets as they only have independent generators $G_I$ (see \cref{def:pZ-mat}),
and thus cannot capture the higher-order polynomial dependencies we observed (\cref{fig:poly-enclosure}).
For this reason, we need to limit the highest degree in~\eqref{eq:process-model-polynomial} to $1$ for the enclosure based on zonotopes,
resulting in much larger enclosures in each layer for zonotopes than for polynomial zonotopes.
An example from the Iris dataset is shown in \cref{fig:iris_results_comparison_zono},
where we are again able to verify the specifications using polynomial zonotopes, but are unable to do so using zonotopes,
as the computed enclosure crosses the decision boundary.
This reinforces the need for a polynomial model to effectively verify process variations.

\begin{figure}[t]
    \centering
    \tikzsetnextfilename{./figures/Results/iris_results_comparison_zono}%
    \definecolor{mycolor1}{rgb}{0.00000,0.44700,0.74100}%
\definecolor{mycolor2}{rgb}{0.85000,0.32500,0.09800}%
\begin{tikzpicture}
\footnotesize
\pgfplotsset{
plotstyle3Z/.style={color=CORAcolorRed, mark=|, forget plot},
plotstyle3PZ/.style={color=CORAcolorBlue, mark=|, forget plot},
plotstyle2/.style={only marks, mark=*, mark options={}, mark size=0.1000pt, legend image post style={mark size=1pt}, draw=black},
plotstyle1Z/.style={area legend, draw=CORAcolorRed, fill=CORAcolorRed, fill opacity=0.2},
plotstyle1PZ/.style={area legend, draw=CORAcolorBlue, fill=CORAcolorBlue, fill opacity=0.2},
plotstyle4/.style={only marks, mark=*, mark options={}, mark size=0.1000pt, legend image post style={mark size=1pt}, draw=black, forget plot},
plotstyle5/.style={draw=CORAcolorSafe,dashed,forget plot}
}
\def\rows{1}
\def\cols{2}
\def\horzsep{1cm}
\def\basepath{./figures/Results/}

\begin{groupplot}[%
group style={rows = \rows, columns = \cols, horizontal sep = \horzsep},
scale only axis,
width=1/\cols*\linewidth -\horzsep,
legend style={legend columns=2,legend to name=legendname, legend cell align=left,/tikz/every even column/.append style={column sep=0.5cm}}
]
\nextgroupplot[xmin=-2,xmax=5,ymin=-1,ymax=3,xlabel={Prediction},ylabel={Label},title={(a)}]
\input{\basepath iris_results_comparison_zono_12.tikz}
\coordinate (top) at (rel axis cs:0,1);
\nextgroupplot[xmin=-1,xmax=5,ymin=-1.5,ymax=3,xlabel={$y_{(0)}$},ylabel={$y_{(1)}$},title={(b)}]
\input{\basepath iris_results_comparison_zono_legends.tikz}
\input{\basepath iris_results_comparison_zono_11.tikz}
\coordinate (bot) at (rel axis cs:1,0);
\end{groupplot}
\path (top|-current bounding box.south)--coordinate(legendpos)(bot|-current bounding box.south);
\node at([yshift=-4ex,xshift=-1ex]legendpos) {\pgfplotslegendfromname{legendname}};

\end{tikzpicture}%
%

    \caption{Enclosure comparison using zonotopes (red) and polynomial zonotopes (ours, blue): (a) projection of all dimensions, and (b) enclosure of the two most significant dimensions.}
    \label{fig:iris_results_comparison_zono}
\end{figure}
\section{Conclusion}\label{sec:conclusion}

We have presented a scalable modeling approach for analog neural networks that incorporates process variations.
Our polynomial-based model is obtained by analyzing the neuron circuit under varying process parameters on the transistor level.
The model is then deployed in a system-level formal verification enclosing the variations in the output of each layer using reachability analysis.
We evaluated our approach on three classification benchmarks with different network architectures, including convolutional neural networks.
In contrast to Monte Carlo simulations, which are not feasible for deeper neural networks (already up to a day for MNIST CNN), our verification approach runs in a few seconds on a normal home computer.
Thus, it can be utilized to optimize the inference circuit design and robustness against process variations, significantly speeding up the design cycle.
In future work, we plan to integrate other sources of variation, such as device mismatch and noise, and extend our approach to larger inference circuits.

\section*{Acknowledgments}
The authors gratefully acknowledge financial support by the project FAI (No. 286525601)
funded by the German Research Foundation (Deutsche Forschungsgemeinschaft, DFG).

%\todo{Max 6 Pages}

\clearpage % References are allowed to be an extra page 7; added \newpage for a better estimate of space left

\balance % balance the reference columns (redundant if the \balance in experiments.tex still lands on page 6)

\bibliographystyle{IEEEtran}                                                    {\footnotesize                                                    
%{\small                                                    
\bibliography{bib}
}

\end{document}